# Video Anomaly Detection Using Pre-Trained Deep Convolutional Neural Nets and Context Mining

Chongke Wu
NSF Center for Cloud and Autonomic Computing
The University of Arizona
Tucson, Arizona
chongkewu@email.arizona.edu

Sicong Shao
NSF Center for Cloud and Autonomic Computing
The University of Arizona
Tucson, Arizona
sicongshao@email.arizona.edu

Cihan Tunc
Department of Computer Science & Engineering
The University of North Texas
Denton, Texas
cihan.tunc@unt.edu

Salim Hariri
NSF Center for Cloud and Autonomic Computing
The University of Arizona
Tucson, Arizona
hariri@ece.arizona.edu

***Abstract*—Anomaly detection is critically important for intelligent surveillance systems to detect in a timely manner any malicious activities. Many video anomaly detection approaches using deep learning methods focus on a single camera video stream with a fixed scenario. These deep learning methods use large-scale training data with large complexity. As a solution, in this paper, we show how to use pre-trained convolutional neural net models to perform feature extraction and context mining, and then use denoising autoencoder with relatively low model complexity to provide efficient and accurate surveillance anomaly detection, which can be useful for the resource-constrained devices such as edge devices of the Internet of Things (IoT). Our anomaly detection model makes decisions based on the high-level features derived from the selected embedded computer vision models such as object classification and object detection. Additionally, we derive contextual properties from the high-level features to further improve the performance of our video anomaly detection method. We use two UCSD datasets to demonstrate that our approach with relatively low model complexity can achieve comparable performance compared to the state-of-the-art approaches.**

## I. INTRODUCTION

Surveillance systems have been widely applied to continuously monitor business operations like banks, retail stores, supermarkets, etc. to improve security and to assist in forensic analysis. Also, there is an increasing trend for using video surveillance in public environments. In 2005, after the installation of the surveillance cameras, the total crime in downtown Baltimore was reduced by 24.85% in four months and in Chicago, violent crime declined one-fifth after the surveillance camera installation [1]. The surveillance systems can be classified as passive monitoring and active monitoring. The former will store the video footage in the system and will be requested as evidence while the latter requires a wireless network connection and enable real-time alert. Although active monitoring can reduce the incident rate and improves public safety, it is resource-intensive and manually intensive operations are required to detect the abnormal events, especially in long videos.

Detecting abnormal events, such as shoplifting and robbery, is a fundamental task of an automated surveillance system. Most video anomaly detection algorithms detect the anomalies by learning from the normal features, which can be classified as low-level features and high-level features [2]. The high-level features provide semantically meaningful activities, though they could have a higher error rate in classification tasks. With the development of the convolutional neural net (CNN)-based computer vision applications, the accuracy of image classification, object detection, and image tracking has achieved better performance compared to the traditional methods [3] [4]. This fact inspires many researchers to use CNNs to extract features [5] [6]. Using high-level features for anomaly detection can reduce model complexity and improve anomaly alert interpretability. However, video anomaly detection studies using features extracted from pre-trained CNNs are still very limited. Hence, it is necessary to investigate the use of anomaly detection framework that integrates multiple pre-trained models and the derived contextual features.

Most deep learning models solve the video anomaly detection by using deep learning models with high model complexity and need large-scale datasets [7]. However, they mostly focus on improving detection performance while paying little attention to model size reduction. However, resource-constrained devices (e.g., edge devices for the IoT platform) are limited by hardware resources such as computing capacity and memory space. Hence, these approaches are not the proper choices for resource-constrained devices. Also, the use of large models may lead to high overhead and poor performance due to the following factors. First, large models have too many hyperparameters, and their performance significantly depends on careful parameter tuning, as in convolutional layer structures [8] [9]. This fact makes the training extremely difficult and time-consuming. Second, it is well known that training of large models needs a large amount of data, and therefore it is hard to be applied to video anomaly detection tasks when only small-scale training data sets are provided. Please note that many real world video anomaly detection tasks still suffer from the issue of insufficient training datasets. Third, large deep learning models used in video anomaly detections are black-box models whose model decisions are hard to interpret.

Inspired by the highly successful techniques that applying pre-trained deep learning models for feature extraction in computer vision tasks, we present a novel approach that uses high-level features from pre-trained CNN models to train the anomaly detection model. This leads to a significant reduction in our anomaly detection model without losing its model interpretability. Further, we integrate context-aware in our video analysis and hence further improvement in detection accuracy and performance. In video analysis, context is used to define the semantics (meaning) of the observed motion and interactions between humans and objects [2]. Hence, we combine the features derived from pre-trained CNNs (such as object position category in background segmentation, multi-object tracking, and object classification) to obtain the context information. Please note that, in this paper, rather than creating individual video models, we focus on using the existing pre-trained models to create an anomaly behavior analysis of the video streams.

The remainder of this paper is organized as follows. In Section II, we discuss the related research on exploring contextual information in video anomaly detection. In Section III, we describe our anomaly surveillance system architecture, the anomaly detection model, as well as the context of high-level features. In Section IV, we present the experimental results of our video anomaly analysis when applied to two University of California San Diego (UCSD) datasets. Finally, we conclude this paper in Section V.

## II. Related work

In this section, we discuss the related work in the commercial field, video anomaly detection methods, and the context-aware based techniques.

Famous video surveillance product suppliers like Hikvision embedded the anomaly detection functionality into their product. In the Safe City Solution of Hikvision, they provide abnormal behavior algorithms in the back-end analysis server, such as detecting sudden running or wandering [11]. Their solution also including face recognition for blacklist alarm. However, they focus more on the specific abnormal detection task. They did not exploit the features for multiple subtasks and could not make self-defined abnormal detection.

Traditional video anomaly detection methods proposed non-deep learning models and use the low-level features, such as probability model with dynamic textures [12] or optical flow[13], Social Force model with grid particle on image [14], and Gaussian Mixture model with compact feature set [15].

The deep learning methods introduce CNNs for feature extraction and autoencoder for anomaly detection [16][17][18]. Base on the CNN and autoencoder, applying Generative Adversary nets (GAN) achieves the state-of-art performance [19][20], while GAN is notoriously computational intensive. Our approach achieves comparable performance while significantly reduces the model complexity by mining the context from the high-level features of pre-trained CNNs.

Analogous to the context-based event recognition methods [10], we classify the context level as image-level and semantic relationship level. At the image level, many low-level event features are extracted to form the appearance context feature or the interaction context feature. Zhu et al. use the histogram of oriented gradient and histogram of optical flow to generate the motion regions and use bag-of-words combined with a multi-class support vector machine (SVM) for anomaly detection [2]. The anomalies are classified into point anomaly, context anomaly, and collective anomaly. At the semantic level, the context captures relationships among the basic event such as the semantic relationships between action, activities, human pose, social role, etc. Zhang et al. use the semantic context information, such as motion pattern and path, to improve abnormal event detection in traffic scenes where an abnormal event is defined as vehicles breaking the traffic rules by considering the trajectories [21]. Pasini et al. present a semantic anomaly detection method to detect anomalies and provides an interpretable explanation [22]. They construct the semantic vector from the textual labels obtained from the pre-trained image labeling software.

## III. Proposed Method

### A. System Design

Most current works of video anomaly detection treat the problem as an independent computer vision task, i.e., they enumerate all the normal observations in the training dataset. Hence, they require a large volume of normal video stream training data, and the model complexity is consequently high [8]. Also, the detection decision is hard to explain since there are no semantic features that can be easily interpreted. They only provide the abnormal event localization, which cannot reflect the temporal causal or the unusual human-object relationship [6]. To overcome these limitations, we integrate the pre-trained CNN-based model that provides the high-level semantic features with the denoising autoencoder-based anomaly detection model. The pre-trained models should have meaningful outputs for visualization and captures the features related to the abnormal event. Our proposed architecture is shown in Fig. 1, where our system is divided into three layers: Hardware layer, middle layer, and application layer.

The hardware layer consists of distributed cameras and related drivers. It transfers the raw video streams into the system's software. The camera position and orientation decides the overall monitoring area and provides the associated coordination of the region of interest. In the hardware layer, there are inherent tasks like camera hand-off and data fusion, which are processed in the next layer. In the middle layer, the raw data fusion module provides organized video streams and the mapping relationship between different cameras. The embedded computer vision tasks in the middle layer then process the video streams and generate the structural data output. The computer vision task is embedded in the middle layer, depending on the application. For example, in a supermarket, the main focus is preventing shoplifting; whereas in a train station, we may use multi-object tracking to provide crowd statistics. Then, the outputs of the selected computer vision tasks are combined and then sent to the application layer for visualization and anomaly detection. The application layer provides the user interface of the surveillance system. Then it includes the basic functionality such as visualizing the video content with the computer vision task results and controlling the camera. The user provides more information for the anomaly behavior based on the defined rules. For instance, in the traffic system, there must be rules that govern the movement of vehicles; for example, when the traffic light color is red, the car should stop. The rules can be implemented as relationships between traffic light colors and cars in the object classification task. The high-level features will be fed into the anomaly detection module that can generate alerts to the user whenever an anomalous event is detected. Furthermore, the sensor tasking module in the application layer receives commands from the user to control the behavior of the cameras like turning and zooming in/out to receive more detailed information on the region of interest.

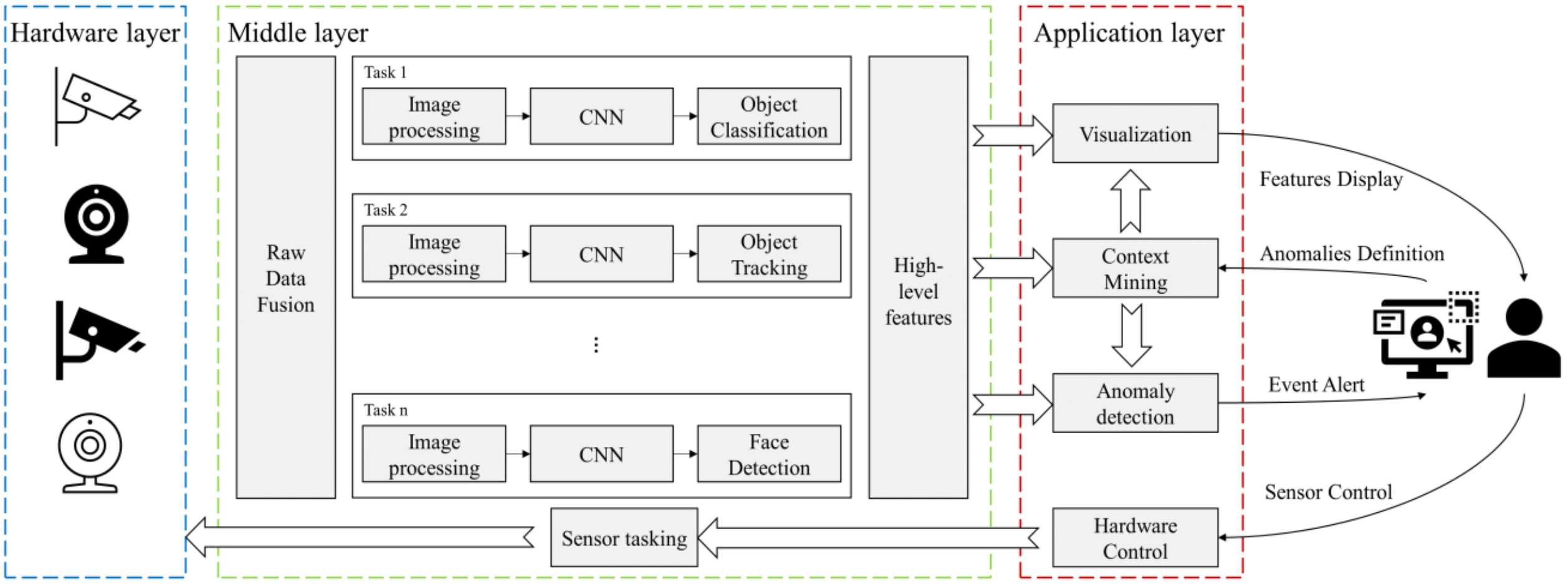

Fig. 1. Design of the proposed intelligent surveillance system. The hardware layer provides the raw data stream to the middle layer. The middle layer embedded with pretrained computer vision modules and output the high-level features to the application layer. In the application layer, the high-level features are visualized and generate the contextual features based on the definition of the anomaly behavior.

### B. *Anomaly Detection Method*

The video surveillance architecture is shown in Fig. 1, we present a novel video anomaly detection method (summarized in Fig. 2). In this method, we process the contextual features (such as human location and background categories relationship) directly from the pre-trained model outputs. In our use case scenario that focuses on crowd surveillance, we choose pre-trained models for background segmentation, object tracking, and object classification. By learning features from pre-trained model output, we focus our research effort on only developing the anomaly detection method rather than studying the individual frames, and this also reduces the complexity of the anomaly detection model.

#### 1) *Feature Extraction*

There are many possible causes of abnormal events, such as abnormal object appearance, abnormal motion, and abnormal object location. We use pre-trained models for background segmentation, object classification, and multi-object tracking to extract the correlated features. To build the background segmentation feature, we consider the Panoptic Feature Pyramid Network (PFPN) [23]. We run this CNN-based model on the Detectron2 platform (The Facebook AI Research software system) [24]. PFPN solves the unified task of instance segmentation and semantic segmentation (for stuff classes: amorphous background regions, e.g., rivers, wall). The model is pre-trained on the COCO train2017 dataset and validated on COCO val2017 [25]. This model has an inference speed of 0.066 seconds per image and masks average precision (AP) of 38.5 on COCO val2017 with GPU V100. The speed allows us to have near real-time (up to 15 FPS) visualization of the background segmentation results. We only select the semantic segmentation for background segmentation. The output can be written as

$$\boldsymbol{L}_{M\times N\times C} = F_{bg}(T) \tag{1}$$

where for the input image at time $T$, the PFPN model $F_{bg}$ outputs a matrix with $C$ classified background labels as well as height $M$ and width $N$. Here we note that this model can be trained on different datasets to improve the segmentation result. For the video anomaly detection task, the background segmentation will only update their results when the vision content changes (e.g., the changing of the ambient light, turning the camera direction, switching the camera). We did not directly utilize the matrix output of background segmentation into the anomaly detection model. Instead, we perform a contextual feature extraction method to process the output and then convert it to a scalar output.

We use the Joint Detection and Embedding (JDE) model [26] to get the pedestrian detection and tracking feature. The JDE model is pre-trained on the MOT-16 training set. The model inference speed is around 38 FPS with the input frame size $576 \times 320$ pixels on an Nvidia Titan Xp GPU. The output is written as:

$$\boldsymbol{p}_i, \boldsymbol{s}_i, \boldsymbol{v}_i = F_{ot}(i, T) \tag{2}$$

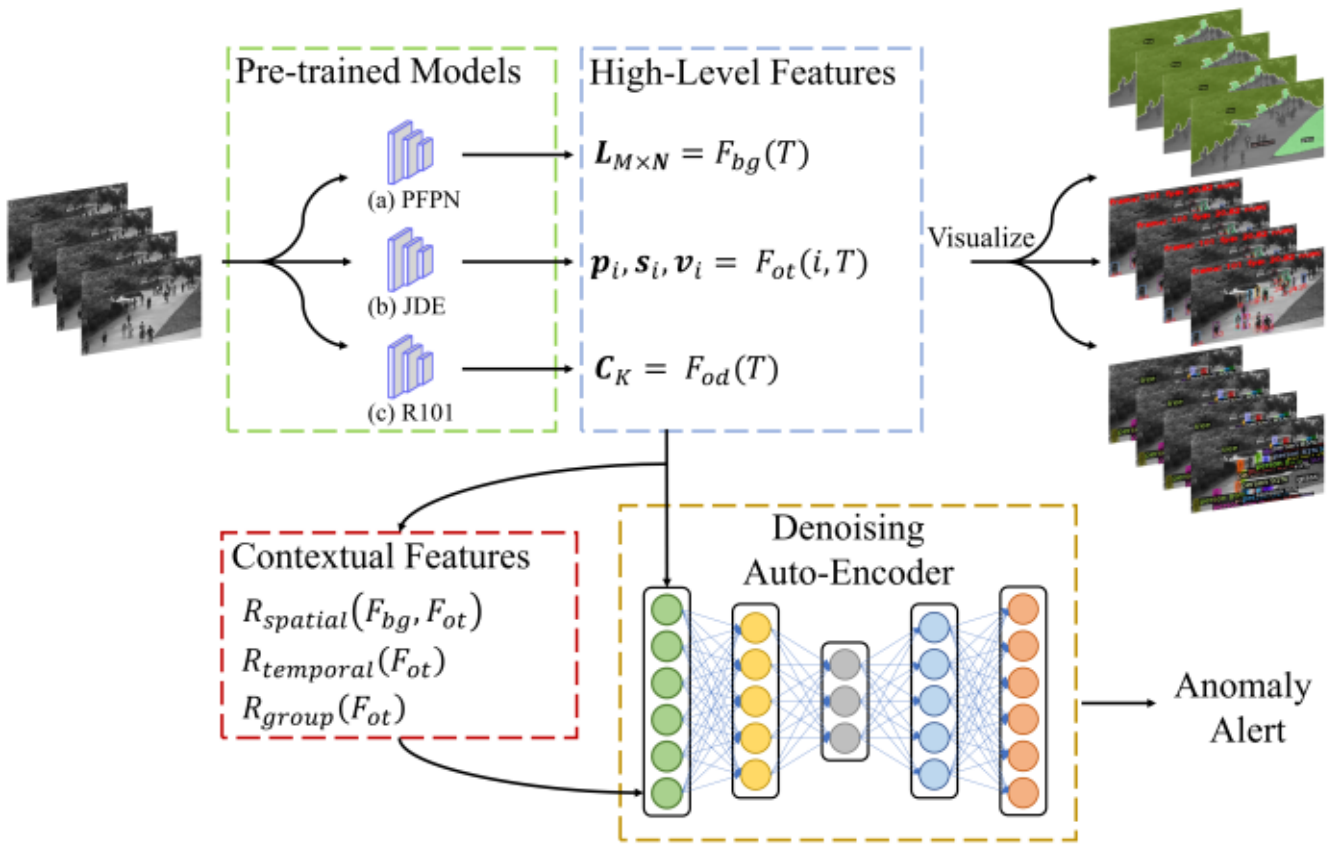

Fig. 2. The architecture of the proposed anomaly detection framework. The computer vision pretrained models generate the high-level features from the video stream. The contextual features mined from the pretrained model output are concatenated with it and feed into the denoising autoencoder.

where $\boldsymbol{p}_i, \boldsymbol{s}_i, \boldsymbol{v}_i$ represents the box coordinates, size (width and height), and the velocity of the person with ID $i$. Given an image at time $T$ as the input of the multiple objects tracking model $F_{ot}$, we will obtain the above outputs for each person. The tracking feature could provide statistical information for each person (trajectories and average speed). We will use these features as the crowd activity analysis in the context mining module.

For the appearance feature, we consider the model ResNet-101 (R101) [27] implemented on the Detectron2 platform. It is pre-trained on the COCO train2017 dataset. The output includes 80 object categories. The R101 model is a CNN-based model that is 101 layers deep. The pre-trained model has an inference speed of 0.051 seconds per image and the box AP of 42.0 on COCO val2017 with GPU V100. The output of the R101 model is written as

$$\boldsymbol{C}_K = F_{od}(T) \tag{3}$$

where $\boldsymbol{C}_K$ is a vector with length equal to the output categories number $K$. When given the frame input at time $T$, the R101 model $F_{od}(T)$ will produce category outputs as a vector. We directly use this vector as an input for the anomaly detection model. We note here the object classification model is crucial to the performance of the video anomaly detection since many abnormal frames are followed with the appearance of the unseen object. We choose the COCO dataset to make it as the baseline for the context mining comparison.

*2) Context Mining*

Even though pre-trained models provide useful features, we still need the inter-relationship between objects. Hence, we process contextual features to improve anomaly detection performance. For that, we classified the extracted contexts as spatial context, temporal context, and group context. The contextual features can reflect prior knowledge from the user who evaluates the pre-trained models' visualization results. If the visualization shows the pre-trained model result is wrong, then the related erroneous context should be adjusted or removed. For example, the user can add a weapon appearance into a blacklist so that an alert should be triggered when a weapon shows up in the video frame. By allowing users to add self-defined contextual features, the searching space for anomaly events can be significantly reduced.

Features that capture the relative spatial relationships among persons or objects of interest are defined as spatial context. We denote the mining spatial relationship process between different pre-trained models result as $R_{spatial}$. The spatial relationship including the intra-spatial relationship and the inter-spatial relationship. The intra-spatial relationship represents the inclusion result $S_1, S_2, \dots, S_C$ of regional classifications $\boldsymbol{L}_{M\times N\times C}$ with height $M$ and width $N$ and the $n$ object detection/tracking results with coordinates $p_i$, $i = 1, \dots, n$. The inter-spatial relationship consists of the adjacent object combinations. One type of spatial anomaly is a certain type of object that is not allowed to appear in a certain type of region. For instance, "trucks are not allowed to drive on the sidewalk". In our case, we use the following formula to represent the spatial relationship between the object tracking and background segmentation:

$$\boldsymbol{O}_{C\cdot J} = R_{spatial}\left(F_{bg}, F_{ot}\right) \tag{4}$$

where the $\boldsymbol{O}_{C\cdot J}$ represents the regional relationship between $C$ types region and $J$ types of tracking objects. The $R_{spatial}\left(F_{bg}, F_{ot}\right)$ denotes considering the intra-spatial relationship $R_{spatial}$ between models $F_{bg}$ and $F_{ot}$. Most work uses the trajectories in training data to determine feasible areas, which means the region without moving objects will be automatically treated as the prohibited region. This kind of mapping has two major shortages. Firstly, it needs to collect enough trajectories in training data to cover the feasible region, which is hard, especially when the monitoring area is large. Secondly, the location will be degenerate when the camera position or orientation is adjusted. By using the spatial relationship between tracking objects and the background type, the above shortages will be overcome since we do not consider the absolute coordinates but the categorized relationship.

Features that capture the relative temporal relationships among the temporal attribute of persons or objects of interest are defined as temporal context. We denote the mining temporal relationship process among the pre-trained models result with timestamps as $R_{temporal}$. The temporal context is widely used in the activity recognition task since the current action could imply the next action. For example, "get off the car" is likely to have "closed-door" behavior followed. In our case, we could consider the speed history of each person then update the Overspeed sign:

$$\boldsymbol{S}_T = R_{temporal}(F_{ot}) \tag{5}$$

where $\boldsymbol{S}_T$ is the frame-level Overspeed sign in the time range $T$. $R_{temporal}(F_{ot})$ denotes the relative relationship $R_{temporal}$ among the results of object tracking output $F_{ot}$. This feature smooths the speed measurement of the object tracking output. In frame-level anomaly detection, the object speed in each frame is not a reliable feature since many movement speeds are periodic (walking, running, riding a bicycle with changing direction, etc.). In this case, we consider the maximal average speed for each person and find the corresponding appearance in each frame.

Finally, we consider mining the group context $R_{group}(F_{ot})$ (frame-level crowd activity statistic) from object tracking features. It includes the min, max, and the median value of the coordinates, and speed. We also use the sum of residuals in the least-squares solution of coordinates and speeds to be the measurement of the crowd sparsity. When all persons are moving in the same direction, then the sum of residuals will equal to zero since the moving direction falls into line (each residual is zero).

*3) Anomaly Detection Method*

For the anomaly detection, we are mainly focusing on the behavior analysis of pedestrians by applying denoising autoencoder (DAE), which is a variant of the basic autoencoder (AE) [28]. DAE is trained through reconstructing a clean input x by a corrupted input $\hat{\mathrm{x}}$, where $\hat{\mathrm{x}} = \mathrm{x} + s \cdot \mathrm{t}$, $s$ is the noise factor, and t is the noise data distribution. In a basic one-layer DAE, The forward propagation for a basic AE with one hidden layer is:

$$\mathrm{h} = f\left(W^{(1)}\hat{\mathrm{x}} + \mathrm{b}_1\right) \tag{6}$$

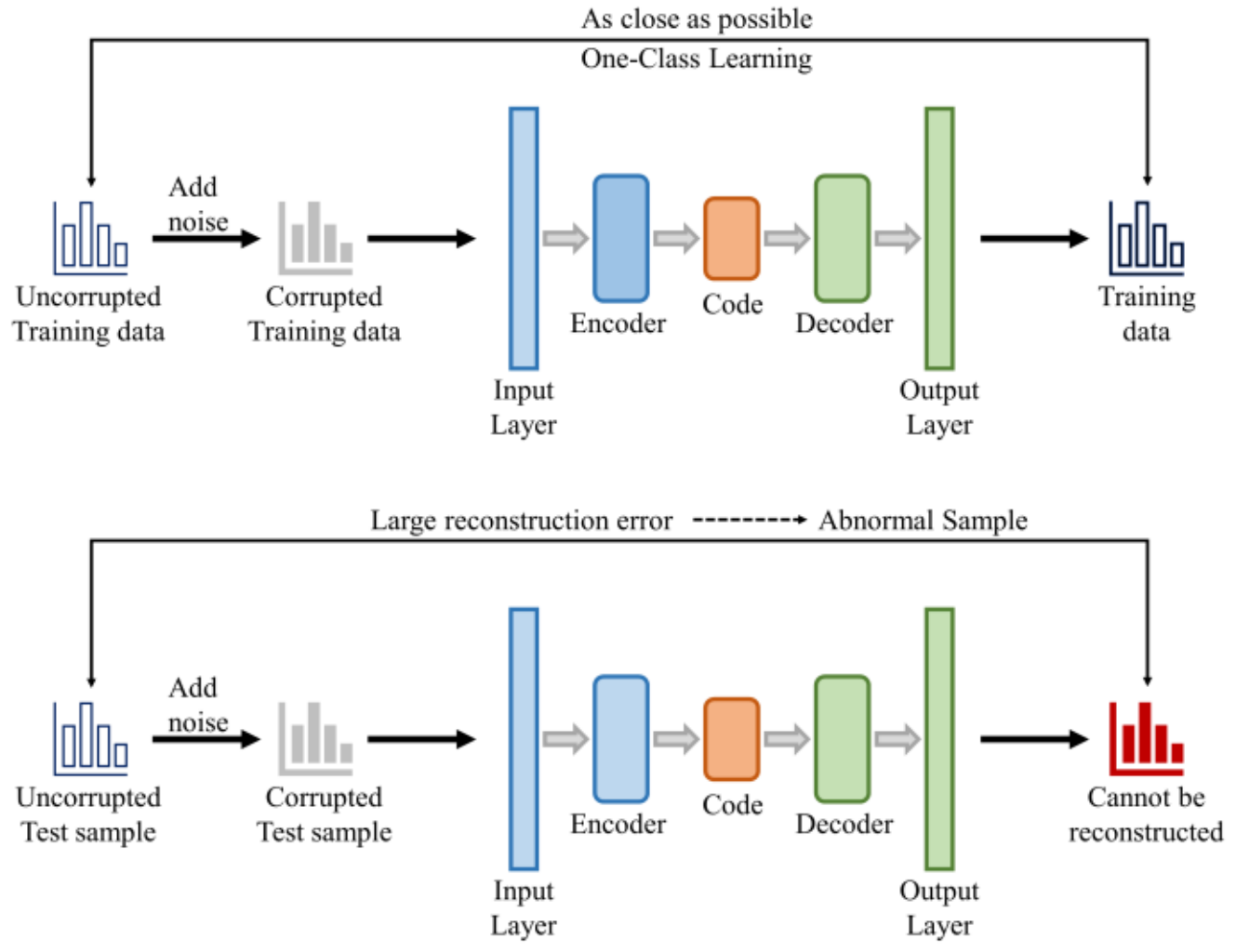

Fig. 3. The architecture of DAE for video anomaly detection.

$$y = f\left(W^{(2)}\mathrm{h} + \mathrm{b}_2\right) \tag{7}$$

where h is the vector of the hidden layer unit activities, y is the reconstruction feature vector in the output layer, $f$ is an activation function, $W^{(1)}$ is the weight matrix between the input layer and the hidden layer, $W^{(2)}$ is the weight matrix between the hidden layer and output layer, and $b_1$ and $b_2$ are the offset vectors. A basic DAE is learned by minimizing the loss function $L(\mathrm{x}, y)$. Deep DAE can be achieved by using multiple hidden layers, that can learn the complicated distribution by given samples due to its multiple feature representation spaces [29]. The backpropagation algorithm [30] is used to train DAE. Our DAE uses the sigmoid activation function for each hidden layer and identity function for the output layer.

One important aspect of our version of DAE is that we use batch normalization (BN) that enables performance improvement and more stable training of DAE [31]. BN uses the mean and variance of batches of training data to perform batch normalization. As a single unit in DAE, its output is given by:

$$y_{NN}(x' : w', b') = g(x'w' + b') \tag{8}$$

where $w'$ is the learned weight, $b'$ is the learned bias, and $x'$ is the input. After applying BN, its output is given by:

$$y_{BN}(x' : w', \gamma, \beta) = g\left(\frac{x'w' - \mu(x'w')}{\sigma(x'w')}\gamma + \beta\right) \tag{9}$$

where $x'$ is a batch training data that can compute the mean $\mu$ and the standard deviation $\sigma$. In the test phase, the parameters $\gamma$ and $\beta$ learned by the original model parameters are used to represent the ranges of inputs to $g$.

Our DAE architecture is shown in Fig. 3. The number of units in the input is determined by the input feature space. To reconstruct observations, the output layer also has the same number of nodes in the input layer. We add three fully connected hidden layers into DAE to form deep DAE. The layer nodes numbers are 50, 30, and 50, respectively (this configuration set provided the best results based on our experiments). The code layer (The middle layer with the 30 nodes) stores the compressed representation space for the input features. Gaussian distribution noise matrix is added into the input vector. Our version of DAE learns the parameters using Adam gradient-based optimization algorithm [32] with mini-batch training to minimize the mean squared error (MSE) used as the reconstruction error. After completing the training phase, our DAE can detect the anomaly. An observation that belongs to normal or abnormal is determined by reconstruction error. During the test phase, an observation is normal if it has a low reconstruction error while it is abnormal if its reconstruction error is large.

## IV. Experimental Results and Evaluation

### A. Dataset

We show the anomaly detection result on the UCSD Ped1 and Ped 2 datasets [12]. The Ped1 dataset has 34 training videos and 36 testing videos. Each video consists of 200 frames with $238 \times 158$ pixels at 30 FPS. The Ped2 dataset has 16 training videos and 12 testing videos. The video frame number of the Ped2 dataset are ranging from 120 to 180 frames with $360 \times 240$ pixels. The training video only includes pedestrians. Both Ped1 and Ped2 provide completed frame-level abnormal labels and partial pixel-level abnormal labels. In this experiment, we only consider the frame-level samples since our work mainly considering the contextual features. The abnormal event includes unexpected entities (bicycle, skateboard, motorcycle, etc.), irregular trajectory (deviate from the major moving direction), and entering the prohibitive region (walking on the grass).

### B. Experiment Setup

We get high-level features from the pre-trained models. The details are demonstrated in Section III. The inference of the pre-trained model is running on the Google Colaboratory [33] server with Tesla V4 GPU.

The DAE is implemented on Tensorflow and Keras. We use Adam optimizer and the MSE loss function to optimize the model. The epoch of training was set up to 200. The batch size was set to 120. In the experiments, we set the noise factor as {0.01, 0.05, 0.1, 0.15, 0.2, 0.25, 0.3, 0.35, 0.4} and choose the better result. The training and test evaluation of the anomaly detection models are running on a computer with the 64-bit Windows 10 Operating System and equipped with 16 GB DDR4 RAM and an Intel Core i7-9750H CPU running at 2.60 GHz.

### C. Visualization

To understand the outputs of context mining, we visualize the results of the embedded computer vision task on both datasets. Fig. 4 and Fig. 5 present examples of the visualization results on both training datasets. For each figure, the images in the first row show the background segmentation results. In the implementation, the user is supposed to select the frames with clear segmentations since their segmentation results are not affected by the ambient light. Only when the camera position is adjusted, the background segmentation should be updated. The images in the second row show the multiple object tracking results. The model assigns a unique ID to each pedestrian. By calculating the difference between the frames, we can get the movement of each person. In the images of the third row, we present the object classification result used as baseline features of our video anomaly detection model, and the accuracy determines the lower bound of our model performance since most anomaly event comes from the occurrence of abnormal objects. When the embedded pre-trained model results are visualized, the user can evaluate the quality of the outputs and decide the principle of formulating contextual features. For instance, if the background segmentation results are unqualified (obvious boundary mismatch or misclassification in pre-trained model evaluation), then we should not consider the relative position context as the anomaly detection features. In our case, we keep all the pre-trained model outputs to generate the contextual features on the Ped1 dataset and we discard the background segmentation results in the Ped2 dataset since the visualization shows that most of the background segmentation results are unsatisfactory. Since we removed the background segmentation, the relevant mined spatial contexts are also removed from the features. In Ped1, the dimension of input features is 100 while in Ped 2 it is 81 since we remove the unreliable features by checking the visualization results.

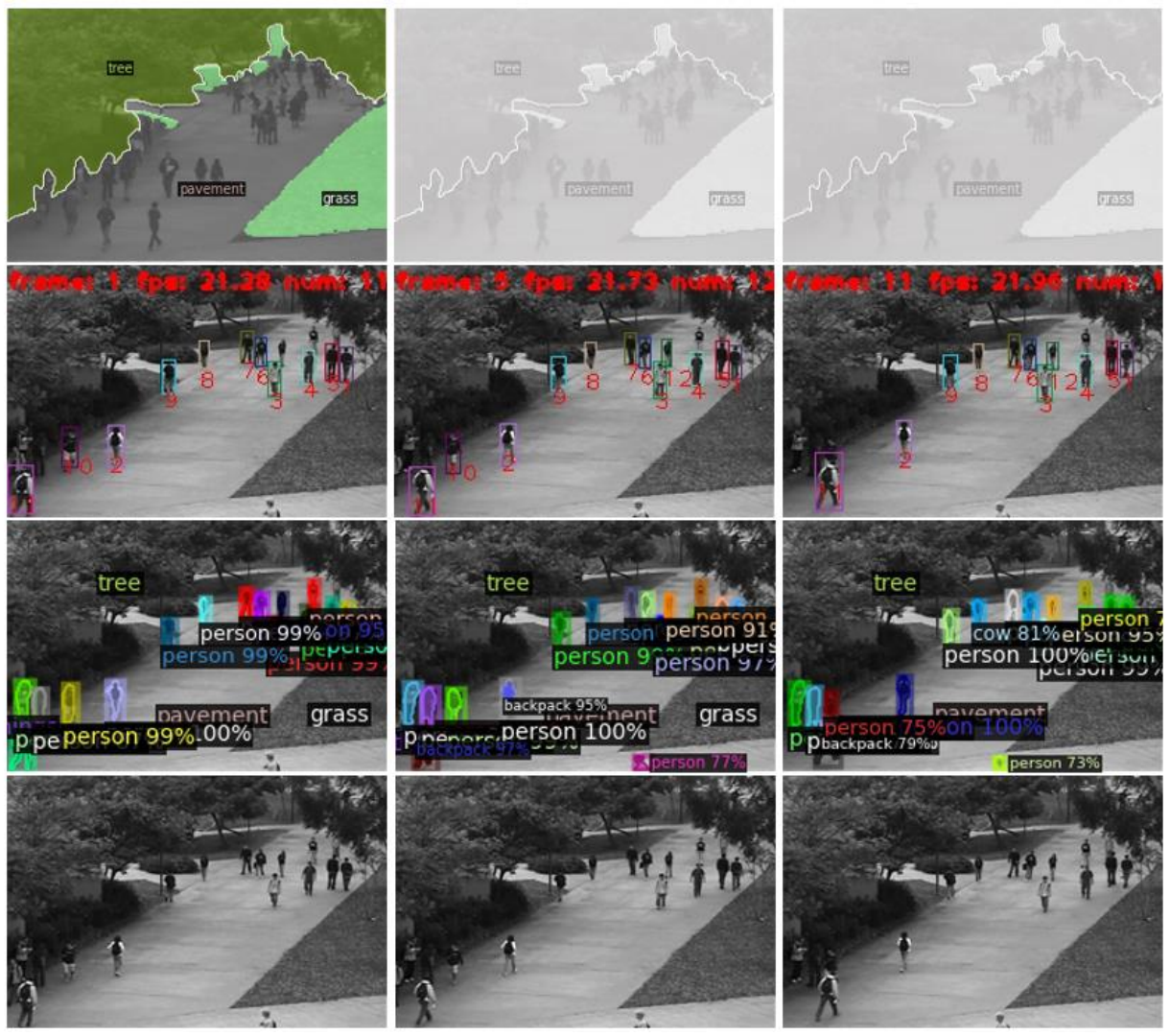

Fig. 4. Visualization of the embedded task result from Ped1 training dataset. From 1st to 4th row is the background segmentation, pedestrian tracking, object classification, origin frames, respectively. Note: the first row is using the same image since the background segmentation should keep constant when the camera is fixed.

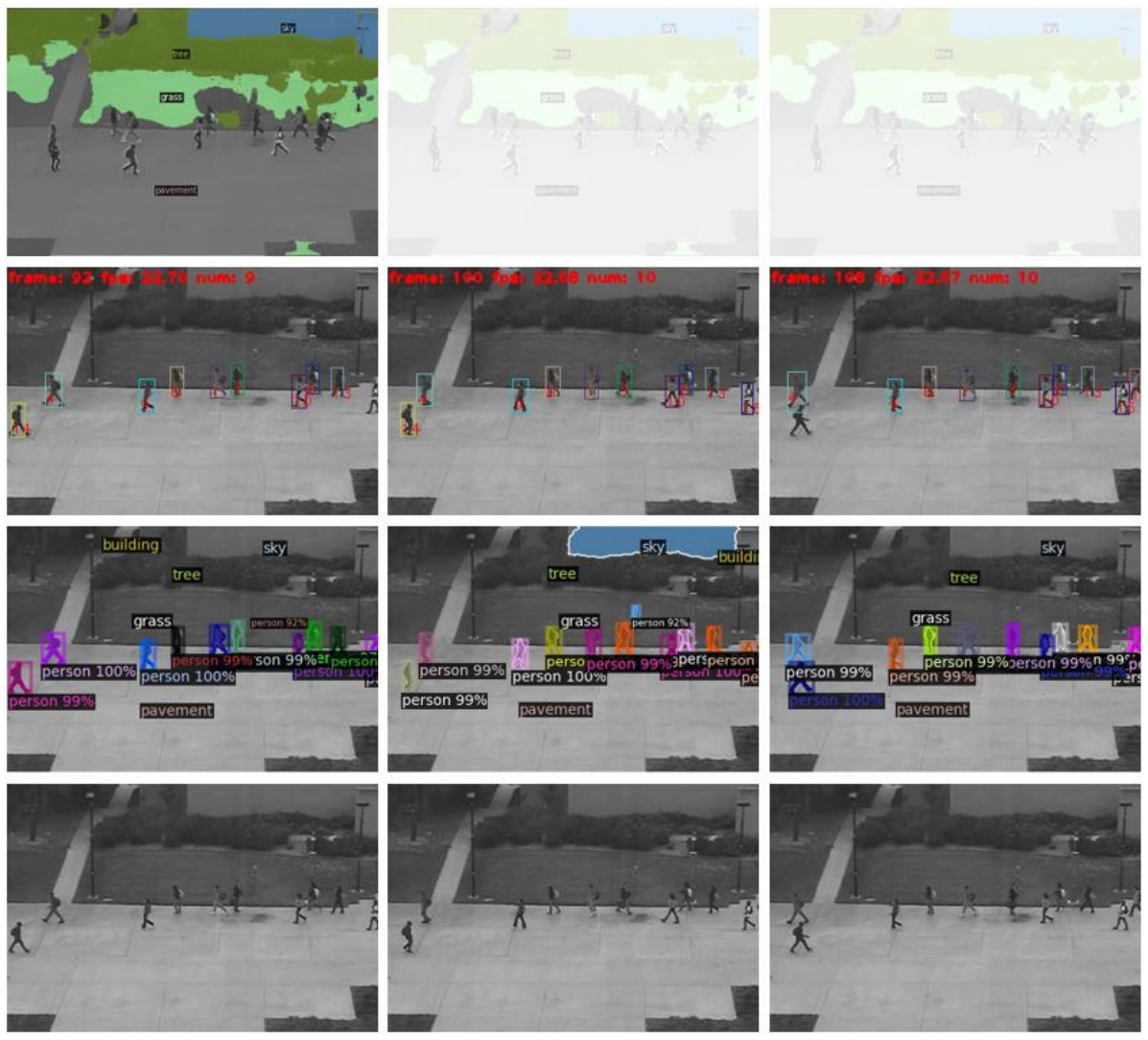

Fig. 5. Visualization of the embedded task result from the Ped2 training dataset. From 1st to 4th row is the background segmentation, pedestrian tracking, object classification, origin frames, respectively. The first row uses the selective fixed results.

## *D. Results*

We evaluate the performance of our video anomaly detection method by considering the effect of the contextual features and training data volume. Receiver operative characteristic curve (ROC curve), area under the ROC curve (AUC), and equal error rate (EER) are the used metrics since they are widely used metrics for Ped1 and Ped2 datasets [34] [12]. To study the effectiveness of our approach, we compare it with state-of-the-art approaches. The ROC curve results are shown in Fig. 7 and Fig. 6. The AUC and EER results are summarized in Table 1.

For the method without contextual features, we only keep the appearance feature (for more information, refer to the approach in [5]). The result shows that the contextual feature effectively integrates the information of movement and semantic result and improves the performance of the anomaly detection method. The AUC is increased by 13.3% in the Ped1 dataset and 7.2% in the Ped2 dataset.

As shown in Table I, our model outperforms the approaches with low complexity (MDT [12], Adam [13], Social force [14], Compact feature set [15], convex polytope ensemble [35], and RBM [36]) and several approaches with large model complexities by adding convolutional layers (ConvAE [8], ConvLSTM-AE [9], Two-Stream R-ConvVAE [16]), and can achieve comparable performance compared to ST-AE [34], and AMDN [17]. Our method achieves the 92.4% AUC on the Ped2 data set and 84.1% AUC on the Ped1 dataset. Hence, the DAE with relatively low model complexity can achieve comparable results using the features derived from the pre-trained deep models. Our model without contextual features achieves 85.2% AUC in Ped2 while 70.8% on Ped1, which means an accurate pre-trained model will improve our final model performance. Most of the competing methods in this study trained the large model while we only consider using the high-level and contextual features derived from pre-trained models to reduce the model complexity for the anomaly detection model. For example, in the Ped1 dataset, the ConvAE model uses the fully convolutional autoencoder [8]. It has 6 convolutional layers and 4 pooling layers in encoder and decoder. The input layer dimension is $238 \times 158 \times 10$. The training process requires up to 16000 epochs to converge. ConvLSTM-AE model adds 10 convolutional long short term memory layers that are interconnected in addition to the convolutional layers [9]. The training process requires up to 60000 epochs. In our case, we only use 3 fully connected layers with an input dimension of 100 and the training process only requires up to 25 epochs to converge in the Ped1 dataset and 200 epochs in the Ped2 dataset with input dimension of 81. We also list the state-of-the-art approaches (STAN [19], ST-CaAE [18], and Optical flow-GAN [20]). In addition to training CNN to learn the spatial features, STAN and Optical flow-GAN takes the Generative Adversarial Network architecture to improve the performance. However, it increases the model complexity. For example, STAN has 17 convolutional layers with kernel size between $5 \times 5$ and $3 \times 3$ where the number of layers has almost tripled compared to ConvAE. ST-CaAE consists of adversarial network ST-AAE and convolutional network ST-CAE. ST-AAE has four 3D convolutional layers and the corresponding four 3D deconvolutional layers, while ST-CAE have three 3D convolutional layers and three 3D deconvolutional layers. Each convolution layer uses kernels with size $3 \times 3 \times 3$, and the number of kernels is 16 in the input convolutional layer. The ST-CaAE also needs to be trained on appearance stream and motion stream, respectively, which further increases the model complexity. Compared to the above models, our approach extracts the complicated part into pre-trained models and only need to train the decision model with the fully connected layers.

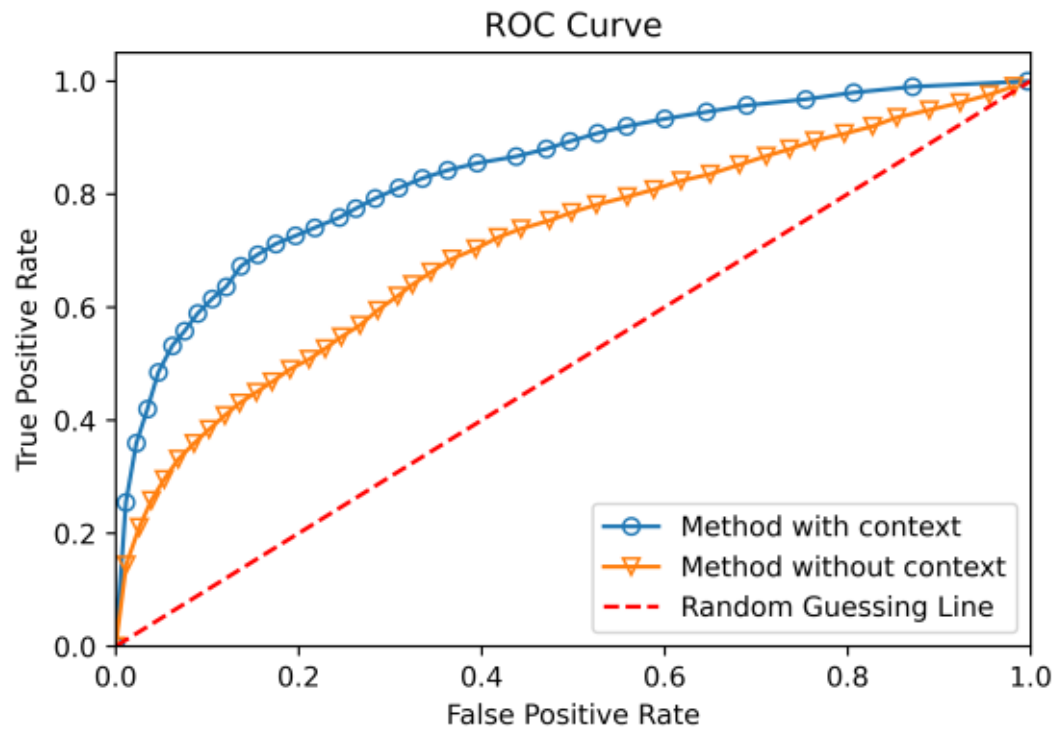

Fig. 7. ROC curve of Ped1 dataset.

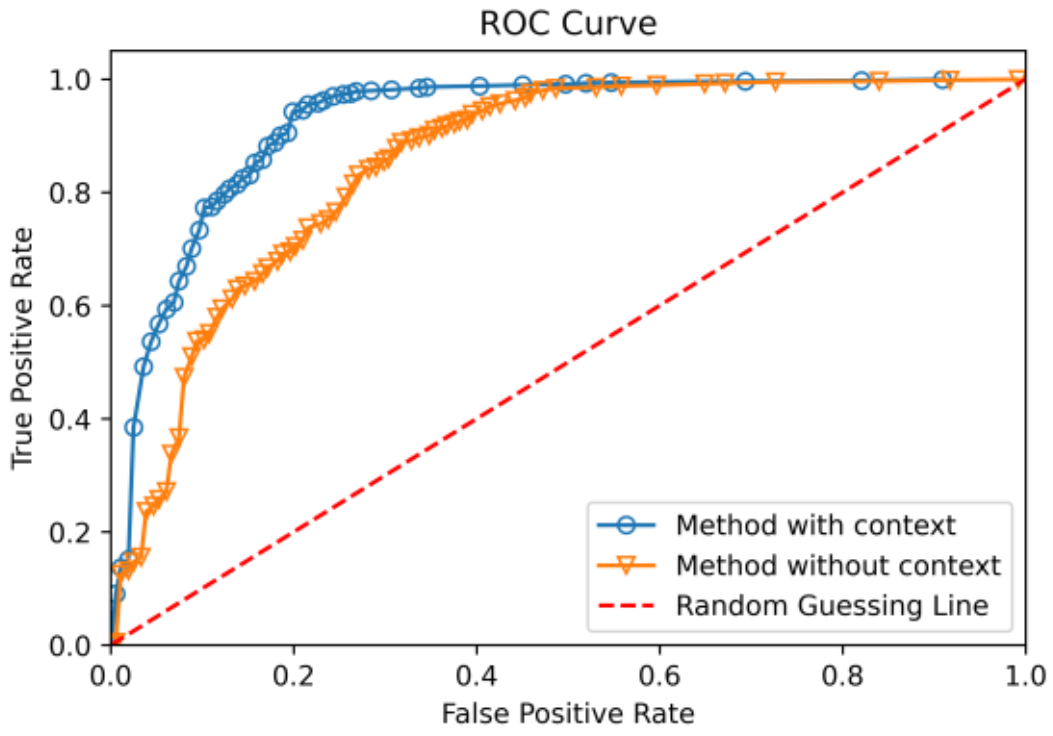

Fig. 6. ROC curve of Ped2 dataset

TABLE I. FRAME-LEVEL PERFORMANCE COMPARISON OF THE ANOMALY EVENT DETECTION

| Methods | Ped1 [12] | | Ped2 [12] | |
|---|---|---|---|---|
| | ***AUC (%)*** | ***EER (%)*** | ***AUC (%)*** | ***EER (%)*** |
| Adam [13] | 65.0 | 38.0 | 63.0 | 42.0 |
| Social force [14] | 67.5 | 31.0 | 63.0 | 42.0 |
| MDT [12] | 81.8 | 25.0 | 82.9 | 25.0 |
| Compact feature set [15] | 82.0 | 21.1 | 84.0 | 19.2 |
| Convex polytope ensemble [35] | 78.2 | 24.0 | 80.7 | 19.0 |
| RBM [36] | 70.3 | 35.4 | 86.4 | 16.5 |
| ST-AE [34] | 89.9 | 12.5 | 87.4 | **12.0** |
| ConvAE [8] | 81.0 | 27.9 | 90.0 | 21.7 |
| ConvLSTM-AE [9] | 75.5 | N/A | 88.1 | N/A |
| Two-Stream R-ConvVAE [16] | 75.0 | 32.4 | 91.7 | 15.5 |
| AMDN [17] | 92.1 | 16.0 | 90.8 | 17.0 |
| STAN [19] | 82.1 | N/A | **96.5** | N/A |
| ST-CaAE [18] | 90.5 | 18.8 | 92.9 | 12.7 |
| Optical flow-GAN [20] | **97.4** | **8** | 93.5 | 14 |
| **Our method** | 84.1 | 23.8 | 92.4 | 14.9 |
| Our method without Context | 70.8 | 35.2 | 85.2 | 24.0 |

Our model also shows the advantages of the interpretability of abnormal event decisions. The other models such as Two-Stream R-ConvVAE use the reconstruction error on each pixel to locate the anomaly region [16]. This method only reflects the spatial features of decision-making and cannot explain the temporal or group anomalies. Since our input features are high-level features and semantically meaningful handcrafted features, we can directly show the reconstruction error vector to explain the decision-making process. Note that here we just use three pre-trained deep models to extract features, and we have shown in the experiments that they are already beneficial, and it is expectable that more profit can be attained by using more pre-trained models that can be used to derive varied features. We leave the possibilities for future exploration.

## V. CONCLUSION

In this work, we have presented a novel design of a video anomaly surveillance system that is based on the high-level features from the pre-trained models and using denoising autoencoder to detect anomalous video events. Two UCSD pedestrian datasets are used to evaluate our approach and to compare it with several state-of-the-art methods. Our experimental results show that contextual features improve model performance. Moreover, our proposed model achieves comparable results while significantly reduce the model complexity and computational overhead of our model. Furthermore, the results produced by our method are easily interpretable. Our approach is not developed to replace state-of-the-art approaches; instead, it offers a better understanding of how pre-trained CNNs can be used for video anomaly detection and provide an alternative approach, especially when training data is not available for large models.

Our method selects three pre-trained models (background segmentation, object classification, and object tracking) to get the appearance feature and spatio-temporal feature. In future work, we aim to improve the performance of our method by including more high-level features such as action recognition feature and key point recognition feature. We also aim to evaluate our approach by using other video datasets.

## Acknowledgment

This work is partly supported by the Air Force Office of Scientific Research (AFOSR) Dynamic Data-Driven Application Systems (DDDAS) award number FA9550-18-1-0427, National Science Foundation (NSF) research projects NSF-1624668 and NSF-1849113, (NSF) DUE-1303362 (Scholarship-for-Service), National Institute of Standards and Technology (NIST) 70NANB18H263, and Department of Energy/National Nuclear Security Administration under Award Number(s) DE-NA0003946.

## References

[1] N.G.L. Vigne, S.S. Lowry, J.A. Markman, and A.M. Dwyer, "Evaluating the use of public surveillance cameras for crime control and prevention." *Washington, DC: US Department of Justice, Office of Community Oriented Policing Services. Urban Institute, Justice Policy Center* (2011).

[2] Y. Zhu, N. M. Nayak, and A. K. Roy-Chowdhury, "Context-Aware Activity Recognition and Anomaly Detection in Video," *IEEE J. Sel. Top. Signal Process.*, vol. 7, no. 1, pp. 91–101, Feb. 2013.

[3] X Gao, S Ram, JJ Rodriguez. "A Post-Processing Scheme for the Performance Improvement of Vehicle Detection in Wide-Area Aerial Imagery." *Signal, Image and Video Processing*, vol. 14, no. 3, pp. 625–633, 2019.

[4] X Gao. "Performance Evaluation of Automatic Object Detection with Post-Processing Schemes under Enhanced Measures in Wide-Area Aerial Imagery." *Multimedia Tools and Applications*, pp. 1-30, 2020.

[5] S. Smeureanu, R.T. Ionescu, M. Popescu, and B. Alexe, "Deep Appearance Features for Abnormal Behavior Detection in Video," *Image Analysis and Processing - ICIAP 2017 Lecture Notes in Computer Science*, pp. 779–789, 2017.

[6] T. N. Nguyen and J. Meunier, "Anomaly Detection in Video Sequence With Appearance-Motion Correspondence," *in 2019 IEEE/CVF International Conference on Computer Vision (ICCV)*, Seoul, Korea (South), Oct. 2019, pp. 1273–1283.

[7] H. Xu, Y. Gao, F. Yu, and T. Darrell, "End-to-End Learning of Driving Models from Large-Scale Video Datasets," *2017 IEEE Conference on Computer Vision and Pattern Recognition (CVPR)*, 2017.

[8] M. Hasan, J. Choi, J. Neumann, A. K. Roy-Chowdhury, and L. S. Davis, "Learning Temporal Regularity in Video Sequences," *2016 IEEE Conference on Computer Vision and Pattern Recognition (CVPR)*, 2016.

[9] W. Luo, W. Liu, and S. Gao, "Remembering history with convolutional LSTM for anomaly detection," *2017 IEEE International Conference on Multimedia and Expo (ICME)*, 2017.

[10] X. Wang and Q. Ji, "Hierarchical Context Modeling for Video Event Recognition," *IEEE Trans. Pattern Anal. Mach. Intell.*, vol. 39, no. 9, pp. 1770–1782, Sep. 2017.

[11] Hikvision.com, 2020. [Online]. Available: https://www.hikvision.com/content/dam/hikvision/en/brochures-download/vertical-solution-brochure/Safe-City-Solution-Brochure.pdf.

[12] V. Mahadevan, W. Li, V. Bhalodia, and N. Vasconcelos, "Anomaly detection in crowded scenes," *2010 IEEE Computer Society Conference on Computer Vision and Pattern Recognition*, 2010.

[13] A. Adam, E. Rivlin, I. Shimshoni, and D. Reinitz, "Robust Real-Time Unusual Event Detection using Multiple Fixed-Location Monitors," *IEEE Transactions on Pattern Analysis and Machine Intelligence*, vol. 30, no. 3, pp. 555–560, Mar. 2008.

[14] R. Mehran, A. Oyama, and M. Shah, "Abnormal crowd behavior detection using social force model," *in IEEE Conference on Computer Vision and Pattern Recognition (CVPR)*, 2009, pp. 935–942.

[15] R. Leyva, V. Sanchez, and C.-T. Li, "Video Anomaly Detection With Compact Feature Sets for Online Performance," *IEEE Transactions on Image Processing*, vol. 26, no. 7, pp. 3463–3478, Jul. 2017.

[16] S. Yan, J. S. Smith, W. Lu, and B. Zhang, "Abnormal Event Detection From Videos Using a Two-Stream Recurrent Variational Autoencoder," *IEEE Transactions on Cognitive and Developmental Systems*, vol. 12, no. 1, pp. 30–42, 2020.

[17] D. Xu, E. Ricci, Y. Yan, J. Song, and N. Sebe, "Learning Deep Representations of Appearance and Motion for Anomalous Event Detection," *Procedings of the British Machine Vision Conference 2015*, pp. 8.1–8.12, 2015.

[18] N. Li, F. Chang, and C. Liu, "Spatial-temporal Cascade Autoencoder for Video Anomaly Detection in Crowded Scenes," *IEEE Transactions on Multimedia*, pp. 1–1, 2020.

[19] S. Lee, H. G. Kim, and Y. M. Ro, "STAN: Spatio-Temporal Adversarial Networks for Abnormal Event Detection," *2018 IEEE International Conference on Acoustics, Speech and Signal Processing (ICASSP)*, pp. 1323-1327, 2018.

[20] M. Ravanbakhsh, M. Nabi, E. Sangineto, L. Marcenaro, C. Regazzoni, and N. Sebe, "Abnormal event detection in videos using generative adversarial nets," *2017 IEEE International Conference on Image Processing (ICIP)*, pp. 1577-1581, 2017.

[21] T. Zhang, S. Liu, C. Xu, and H. Lu, "Mining Semantic Context Information for Intelligent Video Surveillance of Traffic Scenes," *IEEE Trans. Ind. Inf.*, vol. 9, no. 1, pp. 149–160, Feb. 2013.

[22] A. Pasini and E. Baralis, "Detecting Anomalies in Image Classification by Means of Semantic Relationships," *in 2019 IEEE Second International Conference on Artificial Intelligence and Knowledge Engineering (AIKE)*, Sardinia, Italy, Jun. 2019, pp. 231–238.

[23] A. Kirillov, R. Girshick, K. He, and P. Dollar, "Panoptic Feature Pyramid Networks," *in 2019 IEEE Conference on Computer Vision and Pattern Recognition (CVPR)*, Long Beach, CA, USA, pp. 6392–6401, Jun. 2019.

[24] Y. Wu, A. Kirillov, F. Massa, W.Y. Lo, and R. Girshick. "Detectron2." 2019.

[25] T.-Y. Lin, M. Maire, S. Belongie, J. Hays, P. Perona, D. Ramanan, P. Dollár, and C. L. Zitnick, "Microsoft COCO: Common Objects in Context," *Computer Vision – ECCV 2014 Lecture Notes in Computer Science*, pp. 740–755, 2014.

[26] Z. Wang, Z. Liang, Y. Liu, and S. Wang. "Towards Real-Time Multi-Object Tracking." *arXiv preprint* arXiv:1909.12605 (2019).

[27] K. He, X. Zhang, S. Ren, and J. Sun, "Deep Residual Learning for Image Recognition," *in 2016 IEEE Conference on Computer Vision and Pattern Recognition (CVPR)*, Las Vegas, NV, USA, Jun. 2016, pp. 770–778.

[28] Vincent, P. (2011). "A connection between score matching and denoising autoencoders". *Neural computation*, 23(7), 1661-1674.

[29] Lu, Xugang, Yu Tsao, Shigeki Matsuda, and Chiori Hori. "Speech enhancement based on deep denoising autoencoder." *In Interspeech*, vol. 2013, pp. 436-440. 2013.

[30] Y. LeCun, Y. Bengio, and G. Hinton. "Deep learning," *Nature 521*, no. 7553 (2015): 436.

[31] Bjorck, Nils, Carla P. Gomes, Bart Selman, and Kilian Q. Weinberger. "Understanding batch normalization." *In Advances in Neural Information Processing Systems*, pp. 7694-7705. 2018.

[32] Kingma, Diederik P., and Jimmy Ba. "Adam: A method for stochastic optimization." *arXiv preprint* arXiv:1412.6980 (2014).

[33] E. Bisong, "Google Colaboratory," *Building Machine Learning and Deep Learning Models on Google Cloud Platform*, pp. 59–64, 2019.

[34] Y. S. Chong and Y. H. Tay, "Abnormal Event Detection in Videos Using Spatiotemporal Autoencoder," *Advances in Neural Networks - ISNN 2017 Lecture Notes in Computer Science*, pp. 189–196, 2017.

[35] F. Turchini, L. Seidenari, and A. Del Bimbo, "Convex Polytope Ensembles for Spatio-Temporal Anomaly Detection," *in Image Analysis and Processing - ICIAP 2017*. Springer International Publishing, 2017, vol. 10484, pp. 174–184.

[36] H. Vu, D. Phung, T. D. Nguyen, A. Trevors, and S. Venkatesh, "Energy-based Models for Video Anomaly Detection," *arXiv preprint* arXiv:1708.05211, 2017.